%% file: main.tex
\documentclass{article}

\usepackage{PRIMEarxiv}

\usepackage[utf8]{inputenc} 
\usepackage[T1]{fontenc}    
\usepackage{hyperref}       
\usepackage{url}            
\usepackage{booktabs}       
\usepackage{amsfonts}       
\usepackage{amssymb,amsmath,amsthm,array}
\usepackage{nicefrac}       
\usepackage{microtype}      
\usepackage{lipsum}
\usepackage{fancyhdr}       
\usepackage{graphicx}       
\usepackage{color}
\usepackage[square,sort,comma,numbers]{natbib}
\usepackage{subfigure}
\usepackage{multirow}
\usepackage{appendix}

\renewcommand{\thefootnote}{\fnsymbol{footnote}}

\title{Detecting and interpreting faults in vulnerable power grids with machine learning}

\author{
  Odin Foldvik Eikeland \\
  Department of Physics and Technology \\
  UiT-the Arctic University of Norway
   \And
  Inga Setså Holmstrand \\
  Arva Power Company 
   \And
  Sigurd Bakkejord \\
  Arva Power Company 
  \And
  Matteo Chiesa\\
  Department of Physics and Technology \\
  UiT-the Arctic University of Norway 
   \And
  Filippo Maria Bianchi\thanks{Corresponding author: filippo.m.bianchi@uit.no} \\
  Department of Mathematics and Statistics \\
  UiT-the Arctic University of Norway \\
  NORCE Norwegian Research Centre AS
}

\begin{document}
\maketitle
\renewcommand{\thefootnote}{\arabic{footnote}}

\begin{abstract}
Unscheduled power disturbances cause severe consequences both for customers and grid operators. 
To defend against such events, it is necessary to identify the causes of interruptions in the power distribution network. 
In this work, we focus on the power grid of a Norwegian community in the Arctic that experiences several faults whose sources are unknown. 
First, we construct a data set consisting of relevant meteorological data and information about the current power quality logged by power-quality meters. 
Then, we adopt machine-learning techniques to predict the occurrence of faults.
Experimental results show that both linear and non-linear classifiers achieve good classification performance.
This indicates that the considered power-quality and weather variables explain well the power disturbances.
Interpreting the decision process of the classifiers provides valuable insights to understand the main causes of disturbances.
Traditional features selection methods can only indicate which are the variables that, on average, mostly explain the fault occurrences in the dataset.
Besides providing such a global interpretation, it is also important to identify the specific set of variables that explain each individual fault.
To address this challenge, we adopt a recent technique to interpret the decision process of a deep learning model, called Integrated Gradients. The proposed approach allows to gain detailed insights on the occurrence of a specific fault, which are valuable for the distribution system operators to implement strategies to prevent and mitigate power disturbances.

\end{abstract}

\section{Introduction}
\label{sec:intro}
Unscheduled power disturbances cause problems for customers and grid operators as they affect all customers connected to the power network, from single households to large industries \cite{chiaradonna2016analyzing,klinger2014power,meles2020impact,shuai2018review}. 
Power failures might have complex and adverse socio-economic consequences in communities that are heavily reliant on the electricity supply \cite{tully2006human,gopinath2018electricity}. 
The distribution system operator (DSO) is contractually obliged to provide a reliable power supply and to compensate customers affected by power interruptions \cite{lovdata}. 
To meet the expected energy demand, the DSOs must implement management plans that account for the underlying infrastructure.

In this study, we focus on disturbances on a power grid in an Arctic region in Northern Norway, where the energy demand from local food industries has increased greatly. 
The growth in energy demand has resulted in more frequent power disturbances, as the current power grid is operating close to its maximum capacity.
One way to improve the reliability of the power supply is to build a new distribution grid that can handle larger power demand. 
However, this is costly, time-consuming, has a huge environmental impact, and contradicts the vision of better utilizing the current electricity grid infrastructure\footnote{\url{https://www.miljodirektoratet.no/publikasjoner/2020/januar-2020/klimakur2030/}} \cite{Decarbonised_future}. 
An alternative solution is to limit the failures and strengthen only the most vulnerable parts of the grid, but this requires first identifying the factors that trigger power disturbances.

The identification of causing factors of faults in the power grid has proven to be a major challenge for the DSO~\cite{shuai2018review}. 
However, the increased availability of energy-related data makes it possible to exploit advanced data analytics techniques to support the development of strategies for improving the reliability of the power grid \cite{sapountzoglou2020generalizable,8892483,8819403,5985483,8378893,bianchi2015short,bianchi2017recurrent}. 
Recent studies based on statistical data analysis and machine learning (ML), indicated that extreme weather conditions are often an important cause of faults in power grids \cite{perera2020quantifying,sabouhi2019electrical,trakas2019spatial,panteli2015influence,de2019flexibility,owerko2018predicting}. 
However, other factors besides weather could likely affect the power quality.

In this work, we explore a wide spectrum of potential causing factors for power failures. 
We consider explanatory variables relative both to weather and high-resolution power-quality data. 
We adopt ML techniques to detect the power disturbances and to identify the factors that mostly explain the power disturbances.

This paper extends our previous study, which analyzed fault data in the Arctic power grid during year 2020~\cite{eikeland2021uncovering}.
There were important shortcomings in the data used in our previous work:

\begin{enumerate}
    \item The machines of the local industries connected to the power grid are so sensitive to the power quality that they experience failures that are not registered in the failure-reporting system of the DSO. 
    \item The resolution of data in 2020 was too low (1-hour) to understand how power consumption truly affects power quality.
\end{enumerate}

To address these issues, new power quality meters were installed on 19 February 2021 in the power grid under analysis. 
These meters log data every minute and register every small voltage variation.
In addition, they provide detailed information about the power quality in the grid, such as the specific phase where the fault is registered, the magnitude of voltage variation, frequency imbalance, and the amount of flicker. 

\paragraph{Contributions}
First, we build a power faults classification dataset in collaboration with the DSO, by collecting variables that are considered as most relevant in explaining power disturbances.
Then, we train different classifiers, including linear classifiers and a deep learning architecture, to detect an incoming fault from the weather and power-quality variables, registered one minute before the specific fault occurs.
As shown in the experimental results, the classifiers manage to detect most of the power disturbances before their onset, demonstrating that high-resolution data from power quality meters in conjunction with weather data are highly informative.

To gain a better understanding about the relationships between the different variables and the power disturbances, we analyze the decision process of the classifiers.
First, we consider traditional features selection methods, which identify which are the most important variables in the dataset that explain the fault occurrence.
While such an approach gives a global overview of the variables that are, on average, the most informative in the dataset, it does not allow to reason about specific cases.

To address this challenge, we adopt a recent technique to interpret the decision process of a deep learning model, called Integrated Gradients (IG). 
For each individual sample IG assigns to each feature a score, whose magnitude indicates how much the value of such feature contributes to determine the class of the sample.
The proposed methodology shows that the classifiers focuses on heterogeneous sets of features when processing different samples.
This indicates that the occurrence of faults can be explained by multiple different patterns in the weather and power-quality variables.
Our findings are valuable to the DSO for implementing strategies to prevent and mitigate power disturbances.


\section{Related work and studies}
\label{sec:relwork}

There exist a vast amount of literature about the detection of different classes of power quality disturbances, such as deviation in voltage, current, and frequency signals. 
For example, Ref.~\cite{mahela2015critical} provides a comprehensive review of more than 150 research studies between 1986 and 2014 on detection and classification of power quality disturbances. 
In another comprehensive and more recent survey, \cite{mishra2019power} reviewed 242 papers on Power Quality and Classification (PQD\&C) techniques based on digital signal processing and ML. 
The survey performed a comparative assessment on various PQD\&C techniques by considering several criteria, such as type of data used, type of PQ disturbance, and classification accuracy.
 
However, fault detection and classification is a reactive process where models try to classify the fault after it has occurred. 
On the other hand, it is often interesting to identify the causing factors and predict the onset of a power fault. 
A fault prediction model should be able to quantify the likelihood of observing a fault in the next period given a set of conditions described by the explanatory variables in the model. 
Additionally, the identification of causing factors for faults will help the DSO to implement strategies to prevent and mitigate incoming faults.
 
There exist some prior relevant work on identifying causing factors for faults in the power grid. 
The causing factors are often divided into two different categories: i) weather conditions, and ii) other factors such as human-related activities (energy consumption).

\subsection{Weather-related faults}
Harsh and severe weather events are considered to be an important source of faults, and several studies have been conducted to address the impact of such events on power quality. 

Owerko et al. predicted power faults in New York City by monitoring weather conditions \cite{owerko2018predicting}. 
The authors deployed a Graph Neural Network to model the spatial relationships between weather stations and improve the prediction performance.



The impact of seasonal weather on forecasting power disturbances was investigated in \cite{michalowska2020impact}. The authors tested the performance of the proposed models by using two different training sets: seasonal or all-year data. It was shown that, in some cases, the prediction performance of the models improved when the training data is limited to a subset corresponding to a particular meteorological season.



The impact of weather variations and extreme weather events on the resilience of energy systems was investigated in~\cite{perera2020quantifying}. The authors developed a stochastic-robust optimization method to consider both low impact variations and extreme events. The method was applied on 30 cities in Sweden. The results indicated that 16\% drop in power supply reliability is due to extreme weather events.



Other examples of relevant work on weather-related faults can be found in Refs.~\cite{panteli2015influence, sabouhi2019electrical,trakas2019spatial,de2019flexibility}. In addition, several risk assessment studies on the impacts of extreme weather hazards such as earthquakes, thunderstorms, and hurricanes can be found in Refs.~\cite{yang2020quantifying,salman2018probabilistic,salman2017probabilistic,salman2017multihazard,mukherjee2018multi,eskandarpour2016machine}.

The works mentioned so far consider only severe weather events and disregard other factors, such as heavy energy load caused by human-related activities. 
Additionally, many methodologies are tested on synthetic data or on public benchmark datasets, which limits the scope of the evaluation and poses constraints on the data acquisition procedure.

\subsection{Alternative approaches for fault detection}
A methodology to predict power faults by analyzing advanced measurement equipment such as Power Quality Analyzers (PQAs) and Phasor Measurement Units (PMUs.) has been proposed in \cite{hoffmann2019incipient}. The study used real-world measurements from nine PQA nodes in the Norwegian grid to predict incipient interruptions, voltage dips, and earth faults. The authors find incipient interruptions easiest to predict, while earth faults and voltage dips are more challenging to predict.

The authors in \cite{hoiem2020comparative}, compared several ML methods to predict power disturbance events such as voltage dips, ground faults, rapid voltage changes, and power interruptions. The Random Forest models achieved the highest performance and the results indicated that voltage dips and rapid voltage changes were the easiest to predict.

The challenge of detecting back-fed ground-faults has been recently addressed in \cite{abusdal2015utilization}. The authors show that faults can be detected by integrating advanced metering infrastructure with a distribution management system. However, the proposed solution is relevant only for DSOs that adopt the OpenDSS software.


The study in \cite{tyvold2020impact} investigated the possibility of predicting voltage anomalies minutes in advance by using a ML model trained on historical power quality analyzers (PQA) data. The voltage data were collected from 49 measuring locations in the Norwegian power grid. The model attempted to predict voltage anomalies 10 minutes in advance based on the presence of early warning signs in the preceding 50 minutes. It was found that the time passed since the previous fault is a major factor that affects the probability of a new imminent fault.

In \cite{rosenlund2020clustering}, the application of clustering and dimensionality reductions techniques to predict unscheduled events were investigated. First, the authors used several techniques to reduce the dimensionality of the data and to cluster events based on analytical features. Then, the fault events were separated from the normal operating conditions. The findings show promising results when using balanced datasets, while the predictive capability is significantly reduced in unbalanced datasets that, however, often appear in real-world case studies.

Other relevant work on fault detection based on ML techniques can be found in Refs.~\cite{fainti2017three, zhou2017partial,manivinnan2017automatic,khokhar2017new,zyabkina2018feature,shuvro2019predicting}. 
In addition, there are some relevant work that adopts novel ML-techniques for detecting and localizing faults in the power distribution network~\cite{sapountzoglou2020generalizable,chen2019fault,khorasgani2019fault,ferrari2011distributed}. 

This section presented several relevant work in predicting faults by assessing either weather effects or human activities. 
One of the goal of our work is to consider, at the same time, a larger amount of weather variables and electricity-related measures as potential causes of power disturbances. 
A close collaboration with the local DSO has provided us with valuable insights about the relevant variables that should monitored to construct a new classification dataset. 
More importantly, none of the previous work has focused on interpreting the decision process of the classifier, which is key to understand the causes of faults and can provide valuable information to improve the power grid reliability.

\section{Power faults dataset}
\label{sec:dataset}

In this study, we focus on a power grid with a radial structure located in the Arctic.
A detailed description of the grid configuration is deferred to Sect.~\ref{sec:power_grid} in the Supplementary Material.
The grid is subject to frequent power faults, which could be caused by weather factors or by the strain of the infrastructure from a local industry, which dominates the load consumption in the power grid.

We prepared a classification dataset where each sample refers to a period when the grid is operating in normal conditions or to a period preceding a fault, respectively.
Each sample is associated with a feature vector $x \in \mathbb{R}^{12}$ and a label $y \in \{0, 1\}$, indicating the normal condition or the imminent fault, respectively.
The feature vector consists of 6 different energy-related variables and 6 different weather variables, summarized in Tab.~\ref{tab:fault_variables}.
A fault is registered when there is at least a 10\% drop in voltage magnitude.
Further details about faults measurement, what the weather and power variables represent, and how they are collected, is described in Sect.~\ref{sec:dataset_construction} in the Supplementary Material.

\begin{table}[!ht]
\centering
\footnotesize
\setlength\tabcolsep{.9em} 
\caption{Variables analyzed to detect faults in the power grid}
\label{tab:fault_variables}
\begin{tabular}{cc}
\toprule
\textbf{Feature ID} & \textbf{Weather variables}       \\ \hline
1                & Wind speed of gust              \\
2                & Wind direction                  \\
3                & Temperature                     \\
4                & Pressure                        \\
5                & Humidity                        \\
6                & Precipitation                   \\ \hline
 & \textbf{Power variables}         \\ \hline
7                & Difference in Frequency         \\
8                & Difference in Voltage imbalance \\
9                & Difference in Active Power      \\
10               & Minimum Power Factor            \\
11               & Difference in Reactive Power    \\
12               & Flicker                         \\ 
\bottomrule
\end{tabular}
\end{table}

The dataset contains $90$ samples representing reported faults ($y=1$), which occurred in the period between 19.02.2021 to 30.04.2021.
Naturally, the amount of samples associated to normal operating conditions is much larger.
In addition, in normal operating conditions the values $x$ from neighboring hours are very similar to each other.
To limit the amount of class imbalance in the dataset and the redundancy in the over-represented class, we arbitrarily subsampled the non-fault class ($y=0$) by taking 1 sample every 60.
In the final dataset, there are $90$ samples representing a reported fault and $1,647$ samples representing normal operating conditions without any power disturbance.

\section{Methodology}
\label{sec:method}

Our approach consists of two steps.
First, we train a classifier to predict the onset of power faults given the value of the electricity and weather variables.
If we obtain a high classification accuracy, we can conclude that there are strong relationships between the variables, $x$, and the occurrence of faults, $y$.
Then, we use two different techniques to highlight the most informative features identified by the classifiers to solve the task.

In Sect.~\ref{sec:linclass} and \ref{sec:nonlinclass}, we describe which classifiers are considered in this study. 
In Sect.~\ref{sec:ig}, we present an approach for interpreting the decision process of a neural network classifier.

\subsection{Linear classifiers}
\label{sec:linclass}

We consider three different linear classifiers.
The first, is a Ridge regression classifier, which first converts the target values into \{\-1, 1\}\ and then treats the problem as a regression task~\cite{bishop2006pattern}. 
The second model is Logistic regression, which uses a logistic function to approximate the probability of binary classification variable~\cite{bishop2006pattern}. 
The third model is the Linear Support Vector Classification model (LinearSVC), which is a type of Support Vector Machine (SVM)~\cite{boser1992training} endowed with a linear kernel.  

Due to the strong class imbalance, we configure each model to assign a class weight that is inversely proportional to the number of samples in each class.
In this way, errors on the underrepresented class (faults, $y=1$) are penalized much more than errors on the larger class (nominal condition, $y=0$).

One advantage of using linear classifiers is that they construct a decision boundary directly in the input space, which allows to easily interpret the decision process of the classifier.
In particular, the linear models assign a weight $w_i$ to each feature $x_i$ in the input space: the higher $w_i$, the more the values of $x_i$ impact the classification outcome.
Therefore, looking at the magnitude of the weights $w_i$ is a simple strategy to estimate the average importance of the features in the dataset for the classification task.

\subsection{Non-linear classifiers}
\label{sec:nonlinclass}

We consider two non-linear classifiers. 
The first, is non-linear SVC classifier equipped with a radial basis function kernel (RBFSVC). 
As for the linear models, also in this case we used class weights inversely proportional to the class frequency.

\begin{figure}[ht!]
    \centering
    \includegraphics[width=0.5\textwidth]{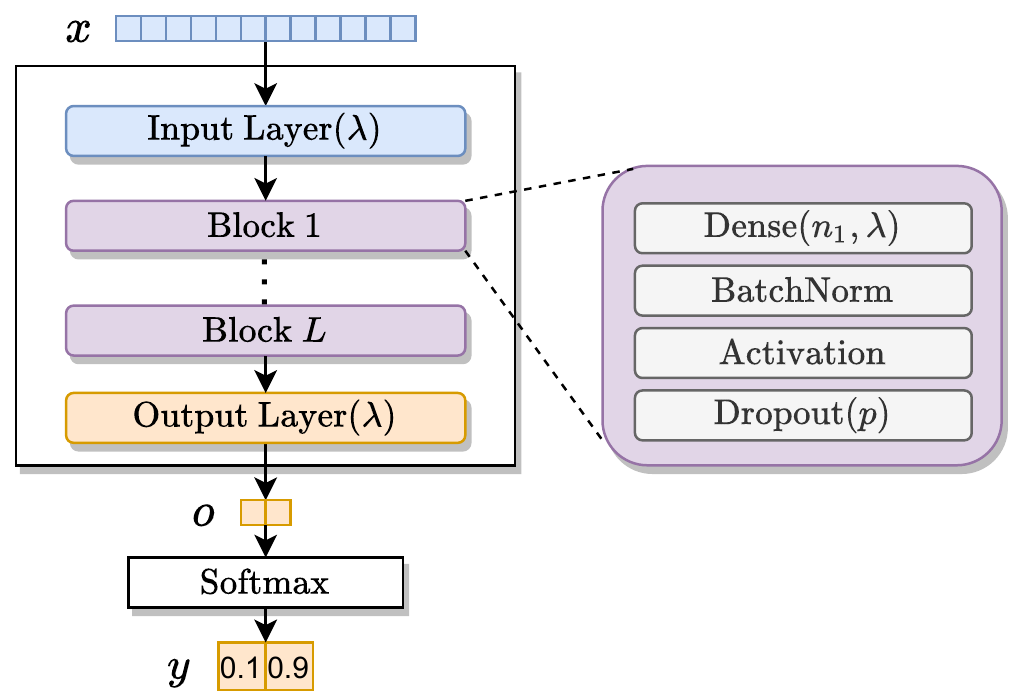}
    \caption{Architecture of the MLP.}
    \label{fig:MLP}
\end{figure}

The second non-linear classifier considered is a Multi-Layer Perceptron (MLP)~\cite{goodfellow2016deep}.
The MLP consists of an input layer that takes input vectors $x \in \mathbb{R}^{12}$, $L$ hidden blocks, an output layer that generates a 2-dimensional output $o \in \mathbb{R}^2$, and a softmax activation that gives the vector of class probabilities $y$.
Each block $l$ consists of a dense layer with $n_l$ units, a Batch Normalization layer~\cite{ioffe2015batch}, a non-linear activation function, and a Dropout layer~\cite{srivastava2014dropout} with dropout probability $p$.
All trainable weights in the MLP, except the biases, are regularized with $\text{L}_2$-norm penalty with strength $\lambda$.
Fig.~\ref{fig:MLP} depicts the MLP architecture.

The MLP is trained by minimizing a cross-entropy loss, using batches of size $b$, and the Adam optimization algorithm~\cite{kingma2014adam} with initial learning rate $r$.
Due to the strong class imbalance in the dataset, we initially trained the MLP by weighting the loss of each sample with a value inversely proportional to the class frequency, like we did for the other classifiers.
However, we found out that the MLP achieved better performance by re-sampling the minority class during training. 
This allows to achieve class balance at the expense of introducing redundancy, by reproposing the same samples multiple times.
We also tried to achieve class balance by subsampling the majority class but, due to the small amount of samples in the fault class, the total number of inputs in each training epoch was too small and the samples from the majority class were shown too few times during training.

\subsection{Interpretation of the MLP results with Integrated Gradients}
\label{sec:ig}

In the following, we introduce the technique adopted to interpret the decision process of the MLP.
A short review of important approaches for interpretability in deep learning, which have been proposed over the past few years (and briefly mentioned hereafter), is deferred to Sect.~\ref{sec:nn_expl} in the Supplementary Material.

Integrated Gradients (IG)~\cite{sundararajan2017axiomatic} is a technique proposed to satisfy two axioms, which are not jointly enforced by other existing attribution schemes (see Sect.~\ref{sec:nn_expl} for details).
According to the first axiom, \textit{sensitivity}, if the input and an \textit{uninformative baseline} differ in exactly one feature, such a feature should be given non-zero attribution.
While interpretability approaches such as LRP~\cite{bach2015pixel} and DeepLiFT~\cite{shrikumar2017learning} ensure sensitivity due to the conservation of total relevance, gradient based methods~\cite{simonyan2013deep, smilkov2017smoothgrad, zeiler2014visualizing, springenberg2014striving} do not guarantee the sensitivity axiom because the saturation at ReLU or MaxPool makes the score function locally ``flat'' with respect to some input features.

The second axiom, \textit{implementation invariance}, states that when two models are functionally equivalent, they must have identical attributions to input features. 
While implementation invariance is mathematically guaranteed by vanilla gradient approaches, the coarse approximation to gradients in LRP and DeepLiFT might break this assumption.

The attribution to feature $i$ given by IG is
\begin{equation}
    \label{eq:ig}
    \text{IG}_i(x) ::= (x_i - x_i') \times \int_{\alpha=0}^1 \frac{\partial F \left( x' + \alpha \times (x -x') \right)}{ \partial x_i} d\alpha,
\end{equation}
where $i$ is an input feature, $x$ is a sample in the dataset, $x'$ is the uninformative baseline, and $\alpha$ is an interpolation constant used to perturb the features of the input sample.
The above definition ensures both the desirable assumptions:
\begin{itemize}
    \item By the Fundamental Theorem of Calculus, IGs sum up to the difference in feature scores and, thus, follow sensitivity;
    \item Since the IG attribution is completely defined in terms of gradients, it ensures implementation invariance.
\end{itemize}

IG has become a popular interpretability technique due to its broad applicability to any differentiable neural network model, ease of implementation, theoretical justifications, and computational efficiency.

\paragraph{Implementation}
IG is a post-hoc explanatory technique that works with any differentiable model, $F(\cdot)$, regardless of its implementation. 
In this paper, we let $F(\cdot)$ be the MLP model described in Section~\ref{sec:nonlinclass} that takes as input tensor the feature vector $x \in \mathbb{R}^{12}$ and generates an output prediction tensor, $o=F(x)$, called \textit{logit}. 
In our case, $o \in \mathbb{R}^{2}$ and $\text{softmax}(o)$ gives the probability of $x$ being ``fault'' and ``non-fault''.

\begin{figure}[ht!]
    \centering
    \includegraphics[width=0.5\textwidth]{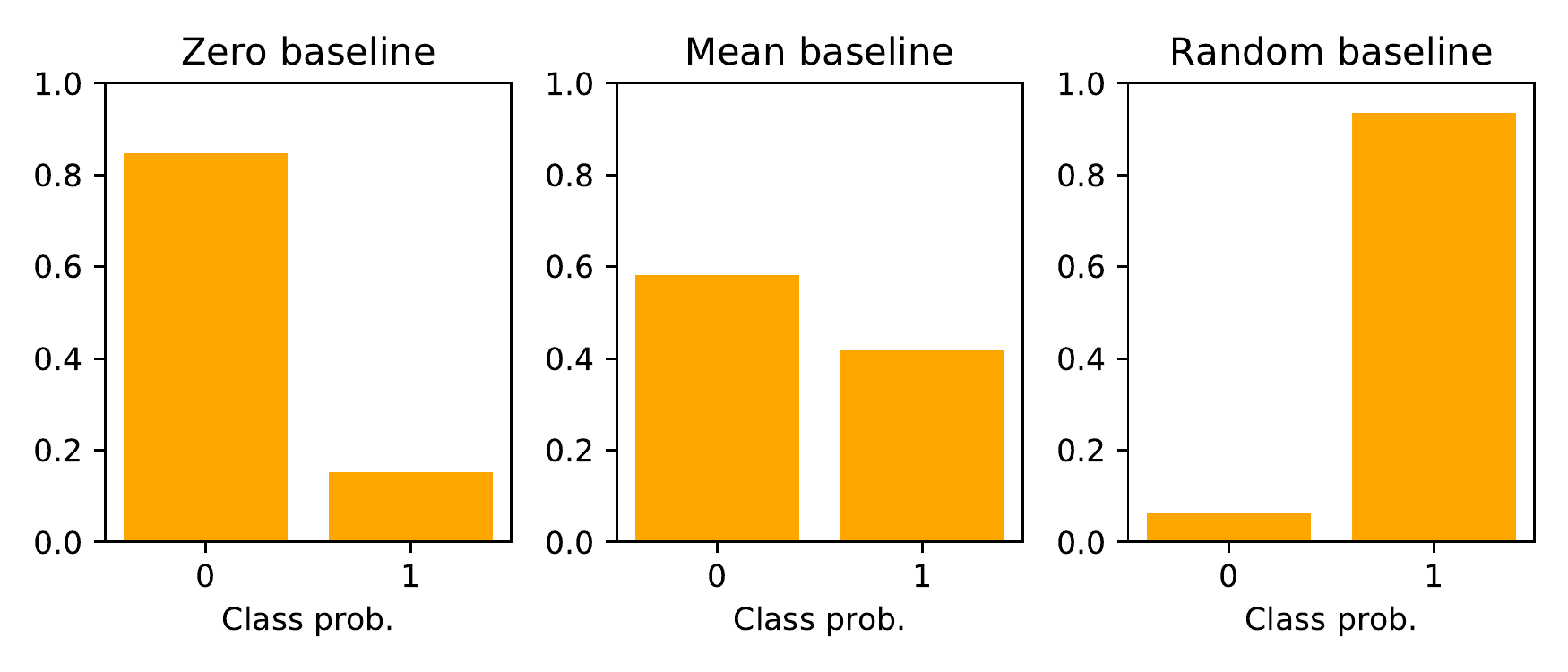}
    \caption{Class probabilities for different baselines on the power-faults dataset.}
    \label{fig:baselines}
\end{figure}

The baseline $x'$ in \eqref{eq:ig} is an uninformative input used as a starting point to compute the IG attributions. 
The baseline is essential to interpret the IG attributions as a function of individual input features.
It is important to choose a baseline that encodes as much as possible the lack of information about the target class $c$. 
In a classification task with multiple classes, we want $\text{softmax}[F(x')]_c \approx 0$. 
In a binary classification task, like in our case, we can chose a baseline that gives equal probability of belonging to both classes, i.e., $\text{softmax}[F(x')]_0 \approx \text{softmax}[F(x')]_1 \approx 0.5$.
In computer vision tasks, a black image (all pixels at 0) is commonly used as a baseline.
However, in our dataset the value 0 might actually be informative because the absence of some specific features can increase the probability of belonging to a specific class (e.g., in the absence of wind it is less likely to observe a fault).
Fig.~\ref{fig:baselines}(left) shows that the MLP assigns with high confidence the zero-baseline $x_z'$ to class $0$ (non-fault).
Therefore, different alternatives should be considered as the baseline.
One option is to cast the binary classification problem into a 3-classes problem and re-train the two that assigns a vector of zeros to a third, dummy class. 
In this way, when using the zero-baseline $x_z'$, we would get $\text{softmax}[F(x_z')]_0 \approx \text{softmax}[F(x_z')]_1 \approx 0$.
Other alternatives are to use a mean-baseline, $x_m'$, which is a vector computed as a weighted average of the features across the two classes or to use, or a random baseline $x_r'$.
In the latter case, the final result is given by averaging the IG attributions computed from several random baselines.
As shown in Figure~\ref{fig:baselines}, the mean baseline gives almost the same probability to classes 0 and 1, while the random baseline has the tendency to assign a strong probability to one of the two classes. 
Therefore, we used the mean baseline in all our experiments.

\begin{figure}[ht!]
    \centering
    \includegraphics[width=0.9\textwidth]{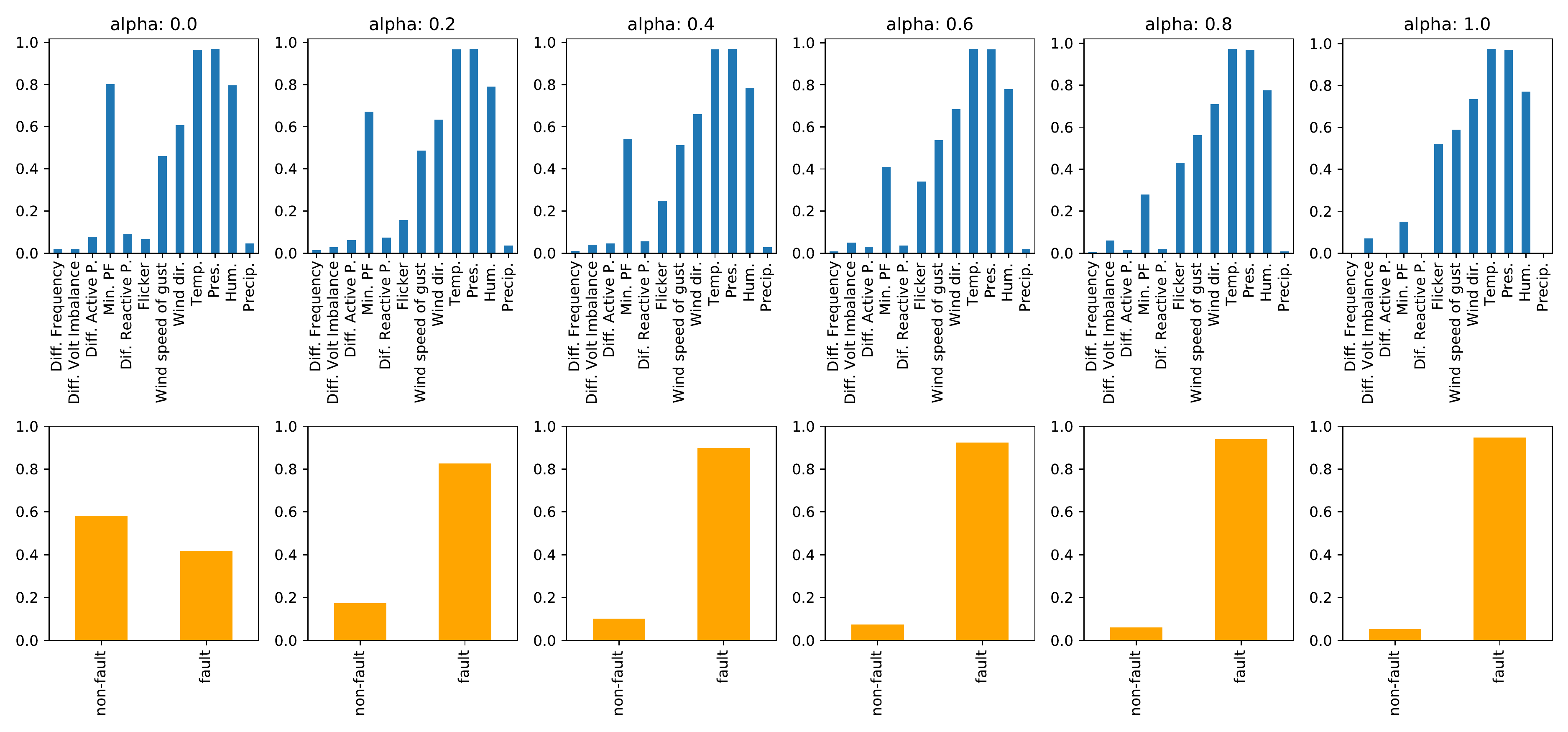}
    \caption{Top row: linear interpolation from the mean-baseline (left) to an actual sample of class fault (right). 
    Bottom row: classification probabilities assigned by the MLP at each step of the interpolation.}
    \label{fig:interp}
\end{figure}

The default path used by the integral in \eqref{eq:ig} is a straight line in the feature space from baseline to the actual input. 
Since the choice of path is inconsequential with respect to the above axioms, we use the straight line path that has the desirable property of being symmetric with respect to both $x$ and $x'$.
The numerical computation of a definite integral is often not tractable and is necessary to resort to numerical approximations.
The Riemann trapezoidal sums offers a good trade-off between accuracy and convergence, and changes \eqref{eq:ig} into:
\begin{equation}
    \label{eq:ig_approx}
    \text{IG}_i^\text{approx}(x) ::= (x_i - x'_i) \times \sum_{k=1}^m \frac{\partial F\left( x' + \frac{k}{m} \times (x -x') \right)}{\partial x_i} \times \frac{1}{m},
\end{equation}
where $m$ is the number of finite steps used to approximate the integral and $\alpha \approx k/m$.
The $m$ samples $\mathcal X = \{ x' + \frac{k}{m} \times (x -x') \}_{k=1}^m$ represent the linear interpolation between the baseline and the input.
Fig.~\ref{fig:interp} depicts such an interpolation path from the mean-baseline to a specific sample of class ``fault'' in our dataset.

\begin{figure}[ht!]
    \centering
    \includegraphics[width=0.6\textwidth]{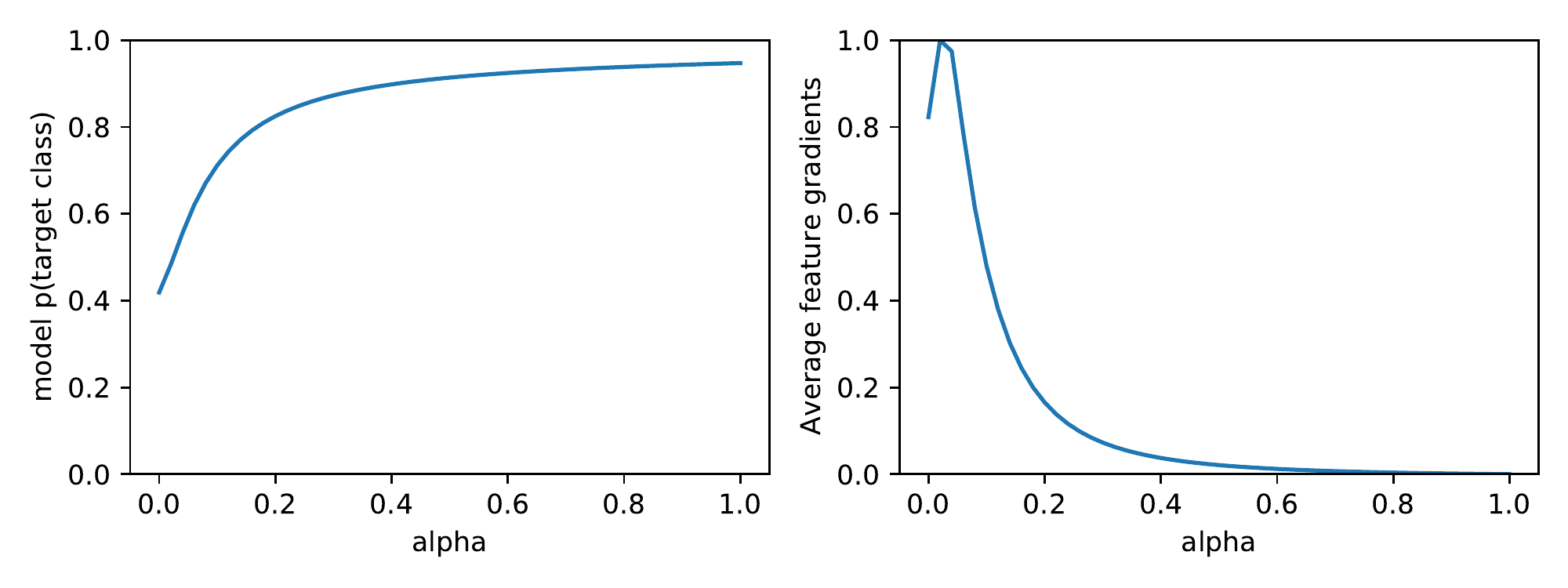}
    \caption{Left: target class predicted probability over $\alpha$. 
    Right: average feature gradients (normalized) over $\alpha$.}
    \label{fig:grads}
\end{figure}

After generating the set of interpolated samples $\mathcal{X}$, we can compute the gradients $\frac{\partial F(\mathcal{X})}{\partial x_i}$ that quantify the relationship between the changes in the input features and the changes in the predictions of the MLP $F$.
Important features will have gradients with steep local slopes with respect to the probability predicted by the model for the target class.
Interestingly, the largest gradient magnitudes generally occur during the first interpolation steps.
This happens because the neural network can saturate, meaning that the magnitude of the local feature gradients can become extremely small and go toward zero resulting in important features having a small gradient. 
Saturation can result in discontinuous feature importances and important features can be missed.
This is the key motivation why rather than simply using the gradients of the actual input, $\frac{\partial F(\mathcal{X})}{\partial x_i}$, IG sums all the gradients accumulated during the whole interpolation path.
This concept is exemplified in Fig.~\ref{fig:grads}(left), showing that the model prediction quickly converges to the correct class in the beginning and then flattens out. 
There could still be less relevant features that the model relies on for correct prediction that differ from the baseline, but the magnitudes of those feature gradients become really small, as shown in Fig.~\ref{fig:grads}(right).
The figure is obtained using the same data of Fig.~\ref{fig:interp}.

\section{Experimental evaluation}
\label{sec:experiments}

After introducing the experimental setting, in Sect.~\ref{sec:exp_comparison} we compare the classification performance of the different classifiers on our dataset. 
Then, in Sect.~\ref{sec:exp_anal} we first analyze the specific samples of class ``fault'' that are missed by the classifiers and, then, we consider two techniques to interpret the decision process of the classifiers. 

\paragraph{Model selection and performance evaluation.}
The linear and the SVM classifiers are implemented with the scikit-learn library\footnote{\url{https://scikit-learn.org/}}, while the MLP is implemented in Tensorflow\footnote{\url{https://www.tensorflow.org/}}.
To evaluate the model performance we first shuffle the data and then perform a stratified k-fold, with $k=5$.
In each fold, $80\%$ of the data are used as training set and the remaining $20\%$ is used as test set.
The training is further divided in two parts: $80\%$ is used to fit the models coefficients and $20\%$ is used as validation set to find the optimal hyperparameters.

The hyperparameters of the linear models and the SVM are optimized with a grid search.
In particular, we optimize the regularization strength in the Ridge regression classifier, Logistic regression, and LinearSVC. 
For the non-linear SVM classifier, we also optimize the width of the radial basis function.

For the MLP, due to the higher amount of hyperparameters and the longer training time, we used the Bayesian optimization strategy implemented in Keras Tuner\footnote{\url{https://keras.io/keras_tuner/}} and evaluated a total of 5,000 configurations. 
In particular, we optimized the number of layers $L$, the number of units $n_l$ in each layer, the $\text{L}_2$ regularization coefficient $\lambda$, the dropout probability $p$, the learning rate $r$, and the type of activation function (ReLU, tanh, or ELU).
We used a fixed batch size $b=32$, an early stopping with patience of 30 epochs, and we reduced the initial learning rate by a factor of $1/2$ when the validation loss was not decreasing for 10 epochs.

Before training the models, the input values $x$ are normalized feature-wise by subtracting the mean and dividing by the standard deviation computed on the training set.
The overall performance of each classification model is the average performance obtained on each test set of the 5 folds.

\paragraph{Performance measures.}
The classification performance is measured by looking at the confusion matrix, which reports the following quantities: 
True Negatives (TN)  -- correctly identified non-faults, 
False Positives (FN) -- non-faults predicted as faults, 
False Negatives (FN) -- faults missed, 
and True Positives (TP) -- faults correctly identified. 
To quantify the performance with a single value we use the F1 score:
\begin{equation}
    F1 = 2 \cdot \frac{\rm{precision}\cdot \rm{recall}}{\rm{precision} + \rm{recall}} = \frac{\rm{TP}}{\rm{TP} + \frac{\rm{FP} + \rm{FN}}{2} }.
\end{equation}
Due to the strong class imbalance in the dataset, we compute a weighted F1 score, i.e., we weight the F1-score obtained for each class by the number of samples in that class and then we compute the average:
\begin{equation}
    F1_{weighted} =\frac{(n_\text{faults}\cdot F1_\text{faults}) + (n_\text{non-faults}\cdot F1_\text{non-faults})}
    {n_\text{faults} + n_\text{non-faults}},
\end{equation}
where $n\_$ and $F1\_$ indicate the number of samples and classification scores for each class, respectively.

\paragraph{Selecting the number of interpolation steps in IG}
The result of the IG attribution depends on the number of steps $m$ (see Eq.~\ref{eq:ig_approx}).
One of the property of IG is completeness, meaning that feature attributions encompass the entire prediction of the model. 
As a consequence, the importance score should capture the individual contribution of each feature to the prediction. 
Therefore, by adding together all the importance scores is possible to recover the entire prediction value for a given sample $x$. 
In particular, we have that the variation in classification score (e.g., the probability of being a fault) is
\[ 
\delta = \sum_i \text{IG}_i(x) - \left( F(x)_c - F(x')_c \right)
\]
where $F(x)_c$ and $F(x')_c$ are the prediction scores for class $c$ when the model takes as input $x$ and $x'$, respectively.
Since we want the $\sum_i \text{IG}_i(x)$ to explain the whole difference in the class attributions, the number of integration steps $m$ should be increased until when $\delta$ becomes as close as possible to zero.
Following this principle, we found $m=100$ to be sufficiently large for our experiments as it gives $\delta < 1e-2$.

\subsection{Classification performance of the different methods}
\label{sec:exp_comparison}

Here, we compare the classification performance obtained by the linear methods, SVM classifier, and the MLP.
The classification performance of each model is reported in Tab.~\ref{tab:Class_scores} in terms of average Weighted F1 score  and the average number of TN, FP, FN, and TP obtained across the 5 folds.
Note that the TN, FP, FN, and TP are rounded to the closest integer.
\begin{table}[!ht]
\centering
\footnotesize
\setlength\tabcolsep{.9em} 
\caption{Classification score for different models}
\label{tab:Class_scores}
\begin{tabular}{lcccccc}
\toprule
& \textbf{Classifier}               & \textbf{TN}                   & \textbf{FP}                   & \textbf{FN}                   & \textbf{TP}                   & \textbf{Weighted F1 score}           \\ \hline
\multicolumn{1}{c}{\multirow{3}{*}{\rotatebox{0}{\textbf{Linear}}} } 
& Ridge Classifier       & 272                  & 57                   & 4                    & 14                   & 0.785                \\
& Logistic regression            & 275                  & 54                   & 5                    & 13                   & 0.756                \\
& LinearSVC                  & 276                  & 54                   & 5                    & 13                   & 0.757                \\
\hline
\multirow{2}{*}{\rotatebox{0}{\textbf{Non-linear}}} 
& RBFSVC                        & 283                  & 46                   & 5                    & 13                   & 0.771                \\
& MLP              & 285                  & 45                   & 4                   & 14                    & 0.803        \\
\bottomrule
\end{tabular}
\end{table}

The MLP classifier achieves top performance with a weighted F1 score of 0.803, followed by the Ridge Classifier and the SVC with rbf kernel that achieve weighted F1 scores 0.785 and 0.771, respectively.
In our case study, is important to miss as few faults as possible, meaning that solutions with very few FN (missed detection) are acceptable even if the number of FP (false alarms) is significant. 
The MLP and Ridge Classifier provide the most promising result with 4 FN and 14 TP on average. 

Finally, it interesting to notice that linear and non-linear models achieve a similar performance. 
This suggests that the two classes are almost linearly separable, i.e., most of the data samples can be separated reasonably well by an
hyper-plane in the input features space. 
On the other hand, the data samples that are misclassified are very entangled and is difficult to find a decision boundary, even if is non-linear, that can correctly separate them. 
The good performance of the classification models motivates the feature interpretation procedure discussed in the next section.

\subsection{Analysis and interpretation of the results}
\label{sec:exp_anal}

For the next analyses, we generate a fixed random train/validation/test split and used the same fold for each model. 
This allows us to analyze in detail the solution obtained by the different methods on a single test set, which contains 18 faults and 330 non-faults.
Interestingly, all models fail to correctly classify as faults the same 5 data samples. 
%
A closer manual investigations on such 5 samples shows the following:

\begin{enumerate}
    \item \textbf{2021-02-22 at 19:29:00:} is an empty measurement, 
    \item \textbf{2021-02-22 at 21:55:00:} is a phase-to-ground fault;
    \item \textbf{2021-02-22 at 22:12:00:} is a phase-to-ground fault;
    \item \textbf{2021-02-26 at 11:58:00:} is an actual fault that was missed by the classifiers;
    \item \textbf{2021-03-02 at 09:29:00:} is a fault with an unusual long duration.
\end{enumerate}

The first FN could have been caused by some type of error, such as a calibration error, in the measurement instruments. 

\begin{figure}[ht!]
    \centering
    \includegraphics[width=0.7\textwidth]{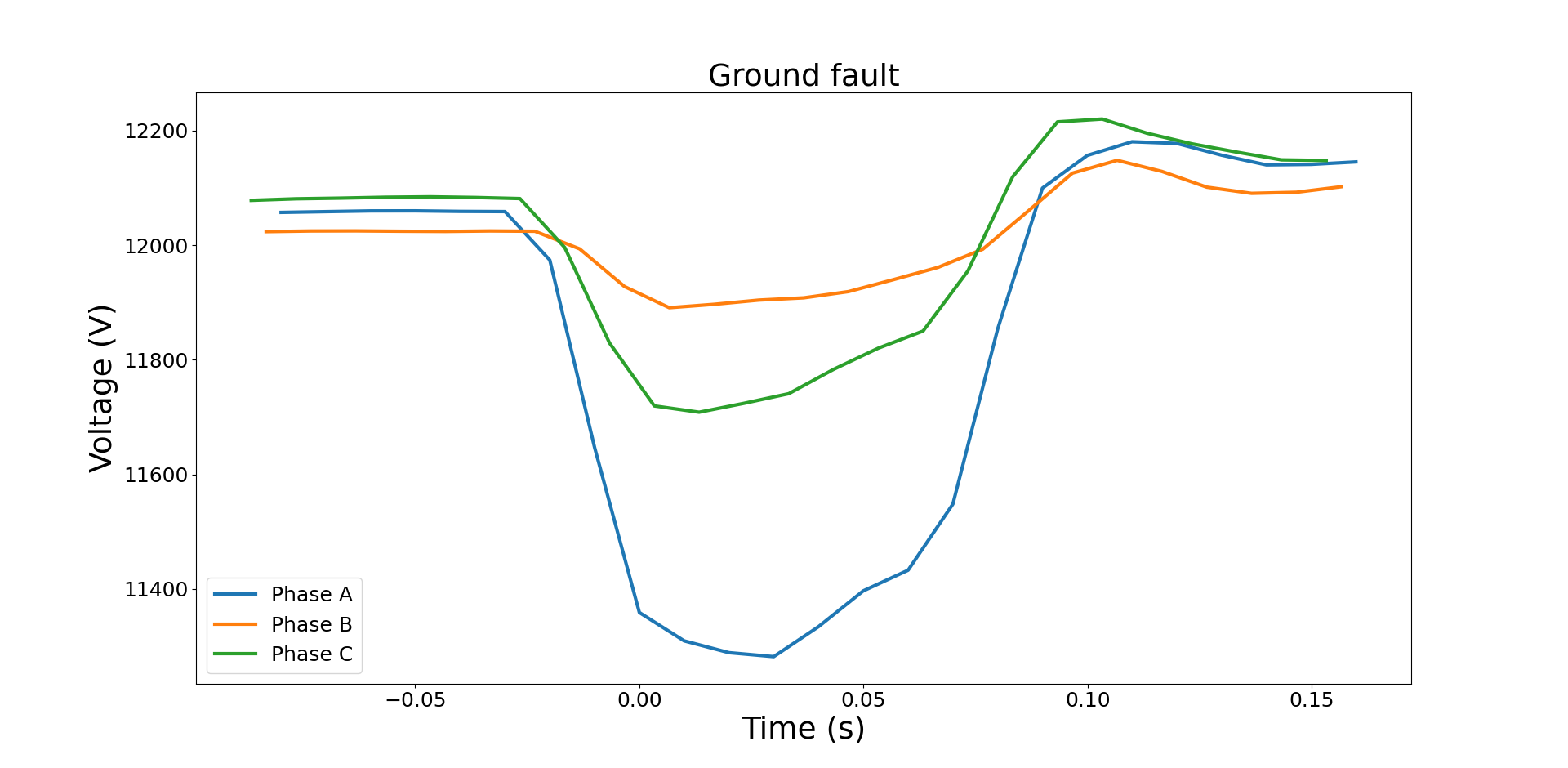}
    \caption{The phases in a phase-to-ground fault incident. The ground fault occurs on Phase A which is decreasing significantly (voltage drop), while the other two in the three-phase system have a minor decrease from nominal voltage level.}
    \label{fig:volt_ground_fault}
\end{figure}

In the case of a ground fault, the electrical transformers connected to the grid break and the power that flows through the transformer flows to the ground. 
When the end of the electrical transformer station that contacts the ground level is on the downstream side, a ground fault occurs \cite{abusdal2015utilization}. 
The ground fault is detected as a reduction of only one of the three phase voltages. 
Fig.~\ref{fig:volt_ground_fault} depicts the phase voltages when the first ground fault occurred: it is possible to see that Phase A decreases significantly, while the other two stay above the nominal voltage value.
It is difficult to explicitly detect ground faults from only weather and electricity load measures considered as input variables, and therefore it is reasonable that the models miss the faults number 2 and 3.

Similarly to the ground faults, the 4\textsuperscript{th} FN could be caused by a factor not described in the weather and electricity variables.
For example, it could have been caused by vegetation or animals interacting with the power lines.

Finally, the 5\textsuperscript{th} FN is a fault which lasts for 200 seconds, while the usual duration of the faults is approximately 25-30 secs.
This suggests that the fault is an anomaly that is not well represented in the dataset and, therefore, is difficult to be classified accurately.

To identify the most important variables that explain the faults, we try to interpret the decision process of the different models.
First, we analyze the coefficients of the linear models, which give a ``global'' interpretation of the variables importance.
Then, we use the IG technique for a ``local'' interpretability of the features that explain the class of a specific data sample.

\begin{figure}[ht!]
    \centering
    \includegraphics[width=0.6\textwidth]{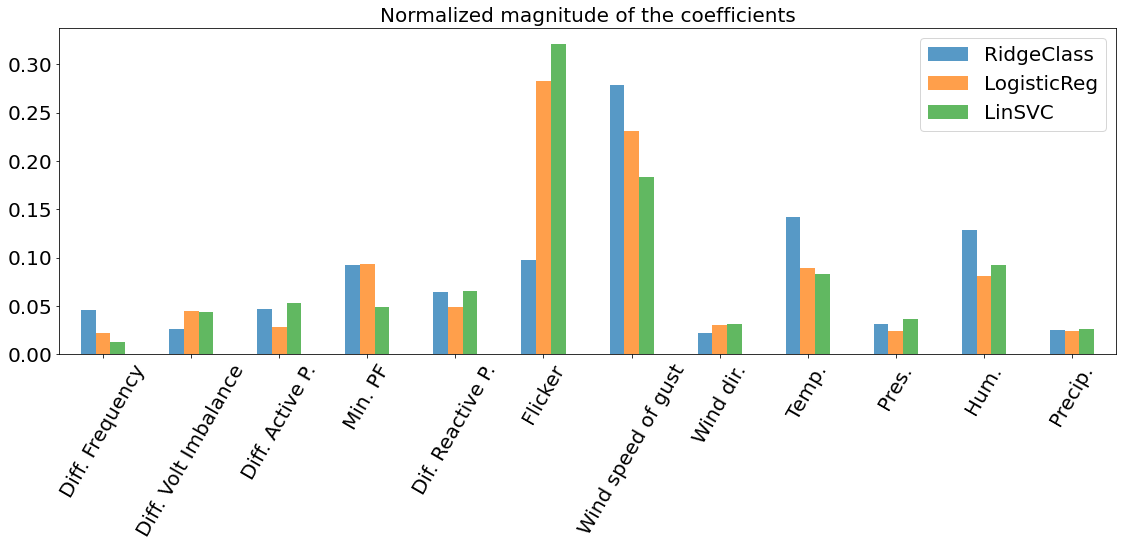}
    \caption{Coefficients' magnitude assigned to each feature by different linear models. High magnitude indicates that the corresponding feature is important.}
    \label{fig:linear_coeff}
\end{figure}

\paragraph{Global interpretability.}
As discussed in Sect.~\ref{sec:linclass}, when using linear models we can interpret the magnitude of the weights assigned to the input features as the global importance of the features for the classification problem. 
Fig.~\ref{fig:linear_coeff} reports the feature weights learned by the three different classifiers.
We observe that in each model the \textit{Wind speed of gust} variable is always associated with a weight with large magnitude.
The Linear SVC and the Logistic Regression classifiers attribute a large importance also to the \textit{Flicker} variable, while the Ridge Regression classifiers weights the other features more uniformly and assigns weights to \textit{Temperature} and \textit{Humidity} that are slightly larger than the weight assigned to \textit{Flicker}.

This analysis suggests that both the industry activity and the weather effects are important in discriminating between the fault and non-fault class. 
According to the linear models, the most important among the power-related variables seems to be \textit{Flicker}, while the \textit{Wind speed of gust} is consistently the most explanatory weather-related variable.
These observations are aligned with experiences of the DSO and the local costumers, as more faults seem to occur when there is high activity at the industries and the machines operates at full load. 
In addition, it has been noted that faults are more likely to occur when there is strong wind, which could create collisions in the cables of the power lines.

\paragraph{Local interpretability.}

The faults correctly classified by the different models are reported in Tab.~\ref{tab:True_pos}, together with the confidence score of the MLP classifier. 
The confidence score can be interpreted as the probability that the MLP believes a sample is a fault.
The MLP correctly classifies with high confidence most fault samples and assigns a probability greater than 90\% to 5 out of 13 samples.
As a side note, the faults do not appear to be clustered around specific days or periods, but they seem to be uniformly distributed over time. 

\begin{table}[!ht]
\centering
\footnotesize
\setlength\tabcolsep{.9em} 
\caption{True positives and confidence score assigned by the MLP classifier.}
\label{tab:True_pos}
\begin{tabular}{ccccc}
\toprule
 & Fault ID &  Confidence (MLP) & Date       & Time     \\ \hline
0   &  31  & 0.71           & 2021-03-02 & 14:30:00 \\
1   &  35  & 0.72           & 2021-02-28 & 20:40:00 \\
2   &  \textbf{52}  & \textbf{0.91}           & 2021-03-24 & 13:39:00 \\
3   &  \textbf{140}  & \textbf{0.92}           & 2021-03-01 & 14:12:00 \\
4   &  163  & 0.60           & 2021-04-07 & 05:16:00 \\
5   &  189 & 0.71           & 2021-03-02 & 14:31:00 \\
6   &  \textbf{227}  & \textbf{0.95}           & 2021-03-01 & 15:04:00 \\
7   &  235  & 0.88           & 2021-02-28 & 14:49:00 \\
8   &  269  & 0.77           & 2021-02-28 & 19:28:00 \\
9   &  271  & 0.86           & 2021-03-24 & 16:01:00 \\
10  &  291  & 0.86           & 2021-03-01 & 14:03:00 \\
11  &  \textbf{304} & \textbf{0.92}           & 2021-03-01 & 13:35:00 \\
12  &  \textbf{316}  & \textbf{0.96}           & 2021-03-24 & 13:38:00 \\
\bottomrule
\end{tabular}
\end{table}

\begin{figure}[ht!]
    \centering
    \begin{minipage}{.45\columnwidth}
    \centering
        \includegraphics[width=0.5\columnwidth]{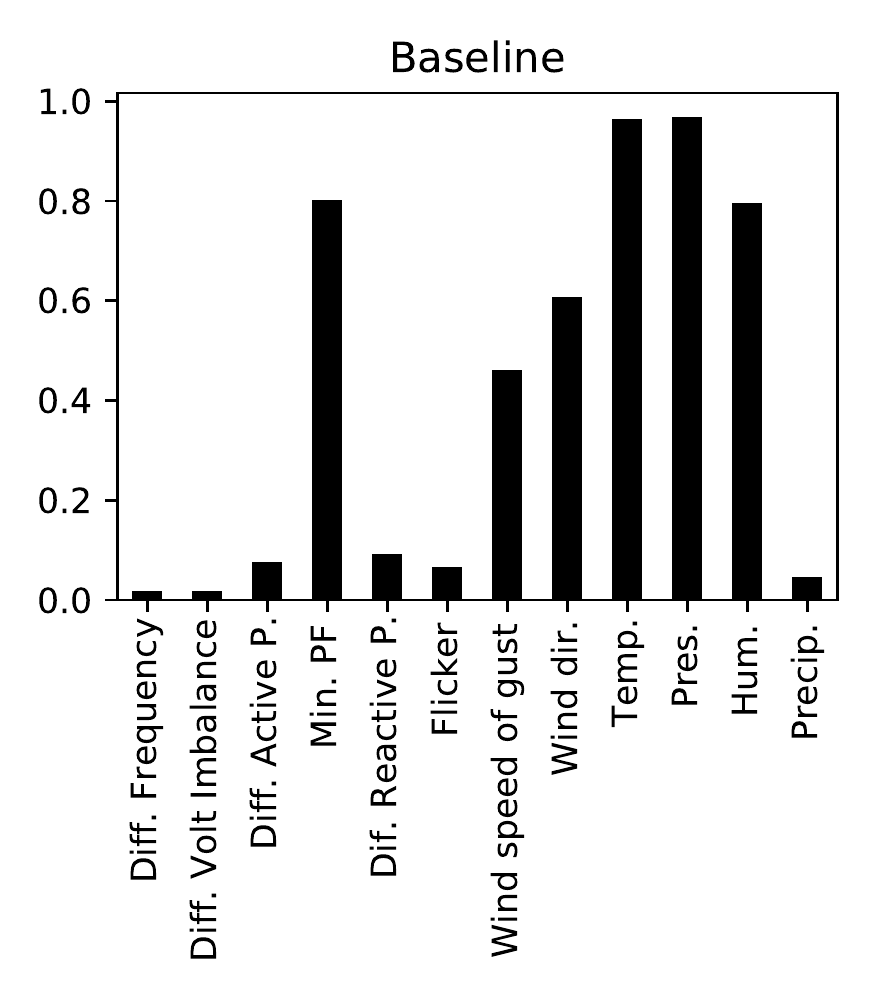}
    \end{minipage}
    ~
    \begin{minipage}{.45\columnwidth}
    \centering
        \includegraphics[width=\columnwidth]{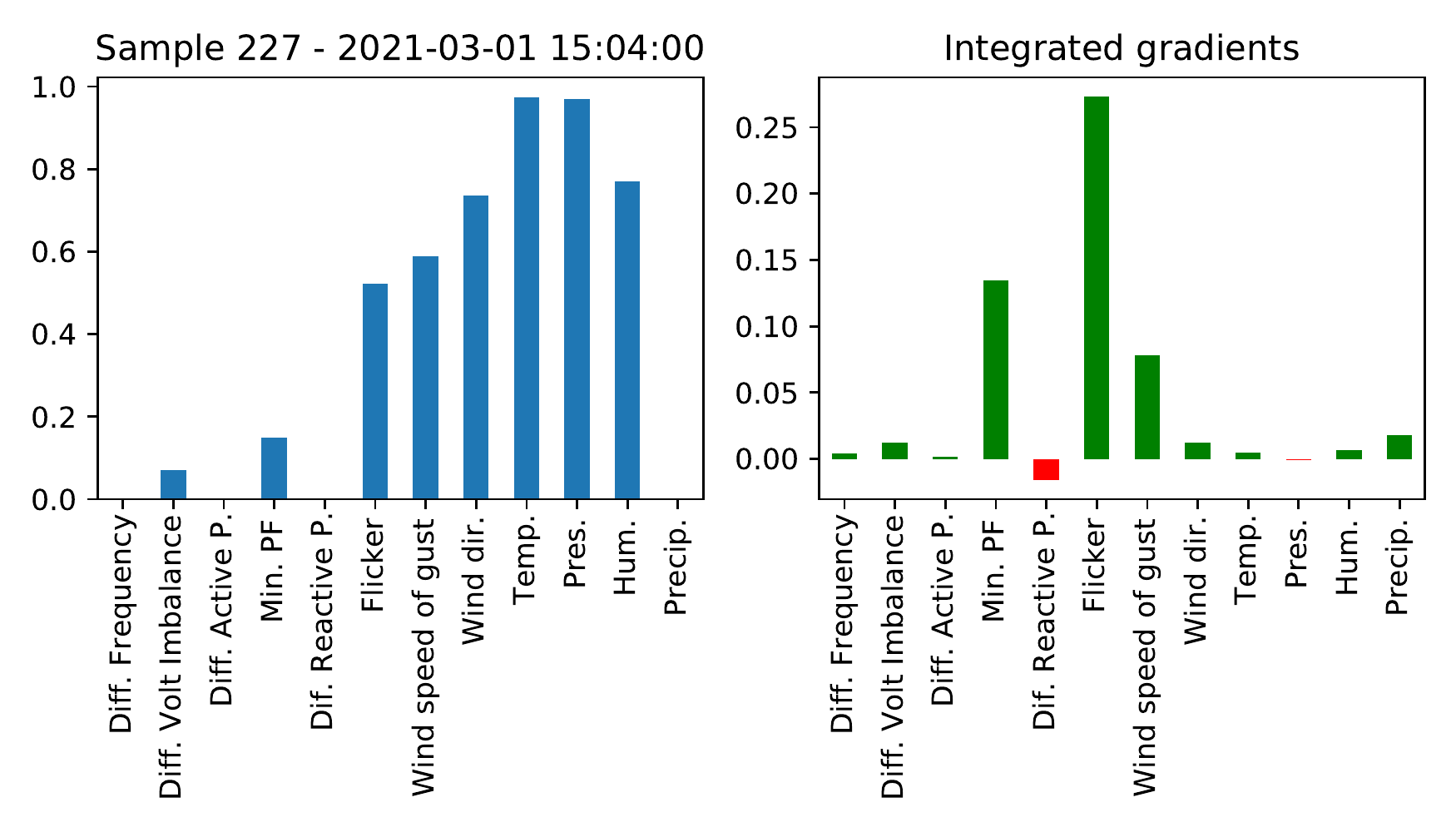}
    \end{minipage}
    
    \begin{minipage}{.45\columnwidth}
    \centering
        \includegraphics[width=\columnwidth]{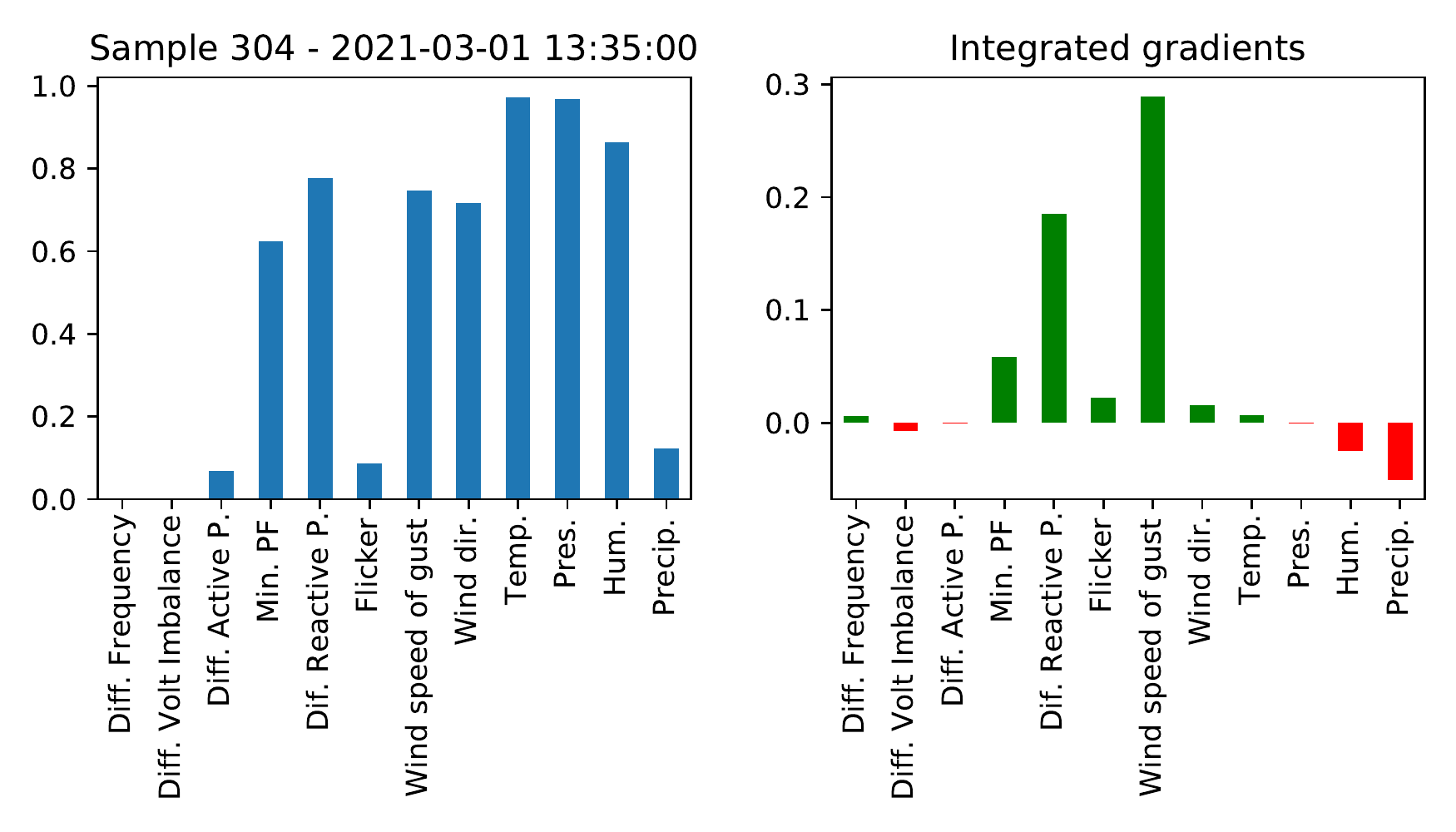}
    \end{minipage}
    ~
    \begin{minipage}{.45\columnwidth}
    \centering
        \includegraphics[width=\columnwidth]{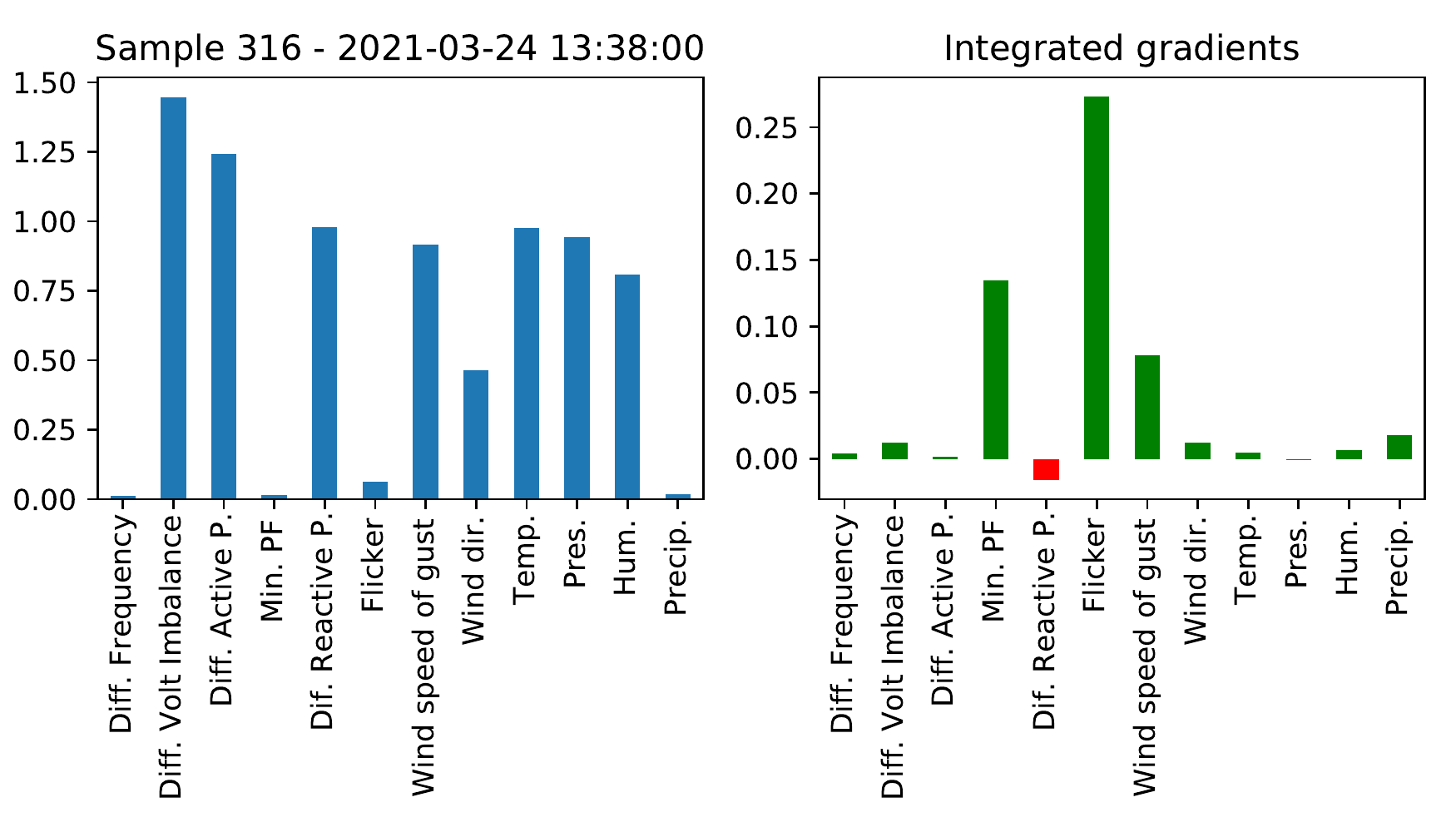}
    \end{minipage}
    
    \begin{minipage}{.45\columnwidth}
    \centering
        \includegraphics[width=\columnwidth]{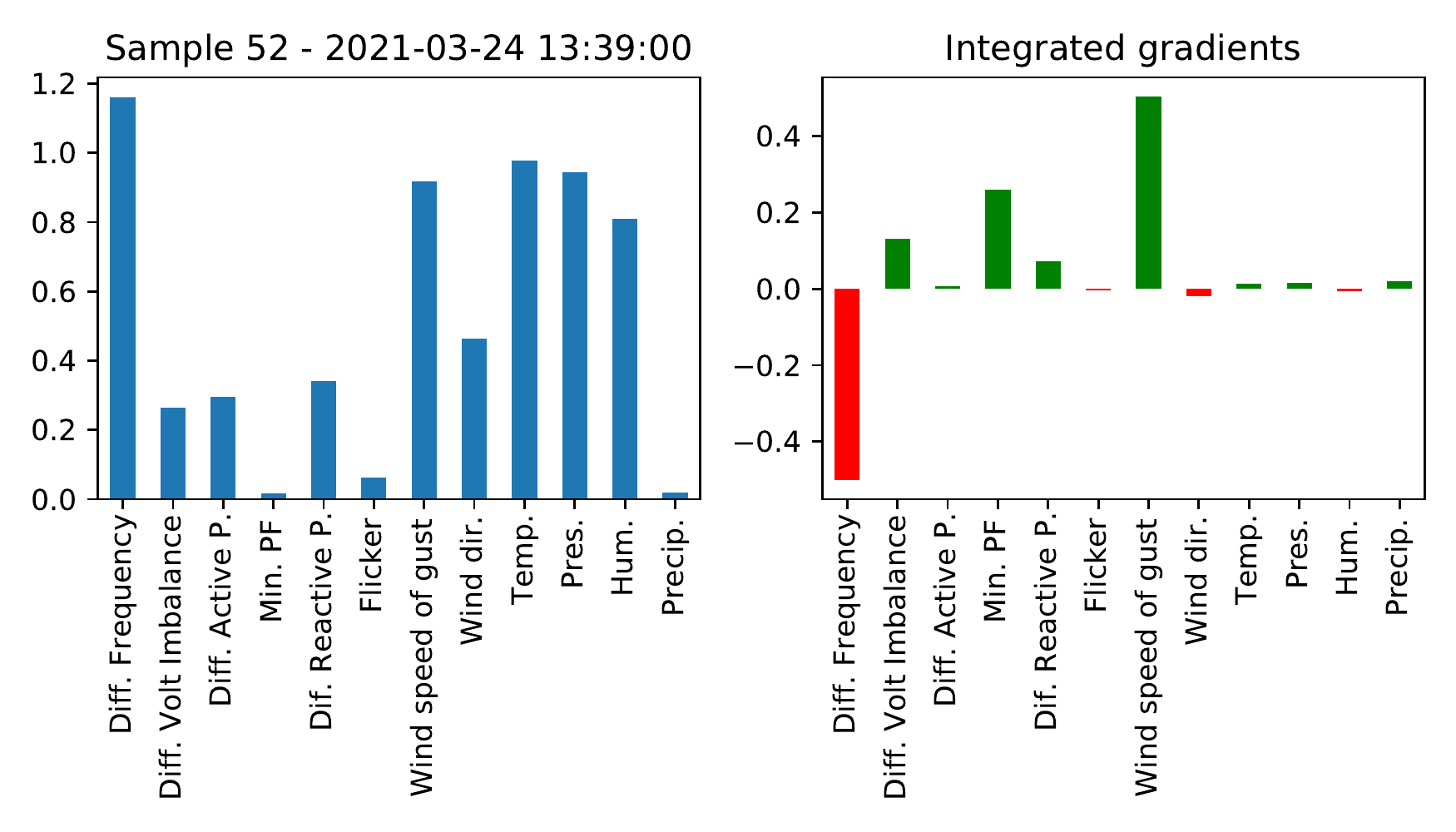}
    \end{minipage}
    ~
    \begin{minipage}{.45\columnwidth}
    \centering
        \includegraphics[width=\columnwidth]{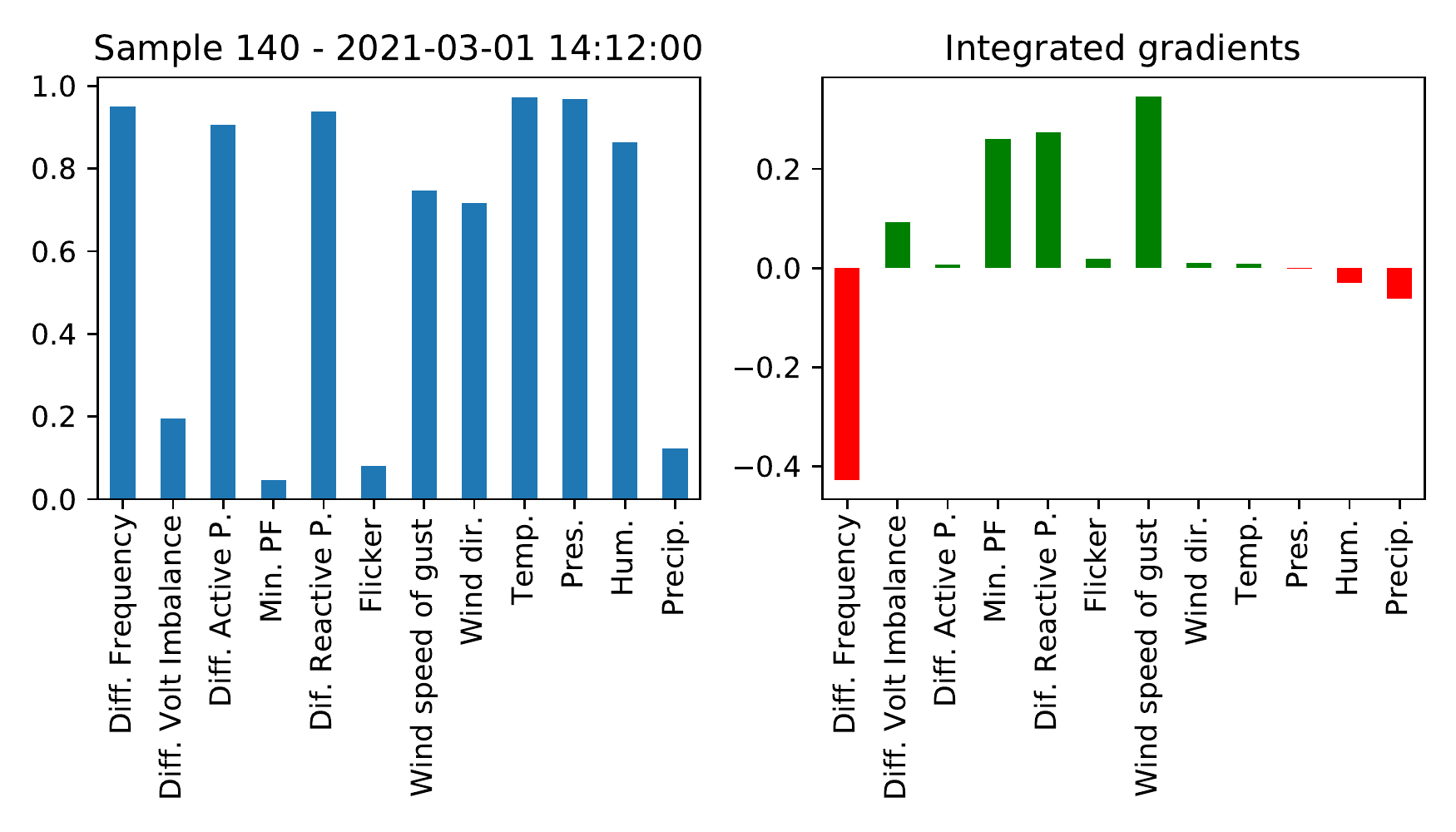}
    \end{minipage}
    \caption{The green bars denote that a feature is important for the classification result. The higher the green bar, the more the feature value in the sample (blue bar) explains the classification result, compared to the value in the baseline (black bar). The red bars means that the value of the features in the sample decrease the confidence of the classifier that the sample is actually a fault.}
    \label{fig:IG_results}
\end{figure}

We focus on the samples 52, 140, 227, 304, and 316 in Tab.~\ref{tab:True_pos}, which are those classified with the highest confidence, and we use IG to identify which are the variables that are most important for the MLP to determine the correct fault class.
The results are reported in Fig.~\ref{fig:IG_results}.
The top-left plot depicts the uninformative baseline, which corresponds to what an ``average'' sample in the dataset looks like. 
The blue bar plots represent the value of the 12 features in the 5 selected samples.
Finally, the green and red bar plots are the output of the IG procedure.

The green bars indicate that a feature is important for the classification result. 
The higher the $i$-th green bar, the more the feature value $x_i$ in the sample (blue bar) explains the classification result, compared to the value $x'_i$ in the baseline (black bar). 
For example, in Sample 227, the value of Flicker is much greater than in the baseline.
IG assigns a high score (tall green bar) to this difference, meaning that the MLP found important the \textit{increment} in Flicker compared to the baseline level for deciding that Sample 227 is a fault. 
Similarly, the MLP found important the \textit{decrement} in Minimum Power Factor compared to the baseline level, to classify Sample 227 as a fault.

A red bar, instead, indicates that a value $x_i$ decreases the confidence in the classifier that the sample is actually a fault, compared to having a baseline value $x'_i$.
For example, the MLP would have been even more confident that Sample 52 and Sample 140 are faults if their Difference in Frequency values would have been as in the baseline. 
In other words, for these two samples the increment of Difference in Frequency is something that decreases the confidence of the classifier that they are faults.

This analysis shows that each sample has different features that are found important by the MLP for the classification.
For example, Sample 227 is classified as a fault mainly because of the above-average value in Flicker; Sample 52 is a fault due to the high value of Wind speed of gust and low value in Minimum Power Factor; for Sample 304 is important that the Difference in Reactive Power is higher than average.

The Minimum Power Factor and Reactive Power are important variables that contribute to explain the current power quality in a power grid. The Power Factor is the ratio of the working power over the apparent power and quantifies the energy efficiency: the lower the power factor, the less efficient is the power usage of the end-customer. 
The Reactive Power is the amount of power dissipated in the system. 
A high amount of reactive power in the system could affect the power quality negatively as there will be less amount of available active power that can be used by the end-customer \cite{mansour2013measurement}. 
Therefore, it is reasonable to observe a relationship between the low value in the Minimum Power factor, and the high Difference in Reactive Power for the fault samples 52 and 304. 

Interestingly, the Minimum Power Factor and Difference in Reactive Power were not emerging as important features with the global interpretability approach, which is based on the weights magnitude of the linear models. 
Indeed, an approach that averages the contribution of the different factors across the whole dataset is likely to conceal the importance of configurations in the features value that appears only in few samples.
On the other hand, by analyzing samples individually, IG can reveal new patterns in the data and allows to gain deeper insights about the true causes underlying specific faults.

\section{Conclusions}
\label{sec:conc}

In this work, we tackled the problem of detecting unscheduled faults in the power grid, which have major consequences for customers, such industries, relying on stable power supply.
In collaboration with the DSO, we built a data set consisting of meteorological and power data variables, which monitor potentially relevant factors to cause power faults. 
Once the dataset was constructed, we trained different classifiers to detect imminent faults from the value of meteorological and power variables.

The classification performance was compared in terms of F1 score and the MLP classifier achieved the top performance, followed by the Ridge Classifier.
The good classification results motivated the interpretation of the decision process learned by the model, as a tool to identify the variables that mostly explain the onset of power faults. 
We explored two different interpretability techniques. 
First, we considered the magnitude of the coefficient of the linear models to quantify the importance that, on average, the different features have to determine if a sample in the dataset is a fault.
The results indicated that the amount of Flicker and Wind speed of gust are the most important variables in explaining the power disturbances.
Such a global interpretability approach averages the contribution of the different factors across the whole dataset and, therefore, might fail to show interesting configurations in the features value that appear only in few samples.

As a second interpretability technique we used the Integrated Gradients to interpret the decision process taken by the MLP classifier on individual samples.
This second approach allowed us to understand what features were considered important to classify a specific sample as a fault.
Interestingly, some samples were classified as faults not only for having high values in Flicker and Wind speed of gust.
In fact, the IG technique showed that the MLP classified as faults samples where the Minimum Power Factor was below average or where the Difference in Reactive Power was higher than average.

The proposed interpretability techniques revealed important patterns in the data, which allow to gain deeper insights about the underlying causes of power faults.
This type of knowledge is fundamental for the DSO operating the grid in our study, which is currently developing strategies for preventing and mitigating incoming faults.
In particular, the local power company is installing a large battery system that should be activated right before an incoming power fault, to supply additional power and avoid instability in the power supply.
Understanding which variable should be monitored to detect an incoming power fault is, therefore, fundamental to implement prevention and mitigation strategies.

\section*{Acknowledgments}
O.F.E, M.C, and F.M.B acknowledge the support from the research project “Transformation to a Renewable \& Smart Rural Power System Community (RENEW)”, connected to the Arctic Centre for Sustainable Energy (ARC) at UiT-the Arctic University of Norway through Grant No. 310026.
We thank the Arva Power Company for providing valuable insight into the problem, and for providing the necessary datasets. We would especially like to thank Inga Setså Holmstrand and Sigurd Bakkejord at Arva Power Company for the valuable collaboration and discussions during the study.

\bibliographystyle{plainnat}  
\bibliography{references}  

\newpage
\input{supplementary.tex}

\end{document}

%% file: supplementary.tex
\appendix
\appendixpage

\section{The investigated power grid}
\label{sec:power_grid}

\begin{figure}[!ht]
    \centering
    \includegraphics[width=.5\columnwidth]{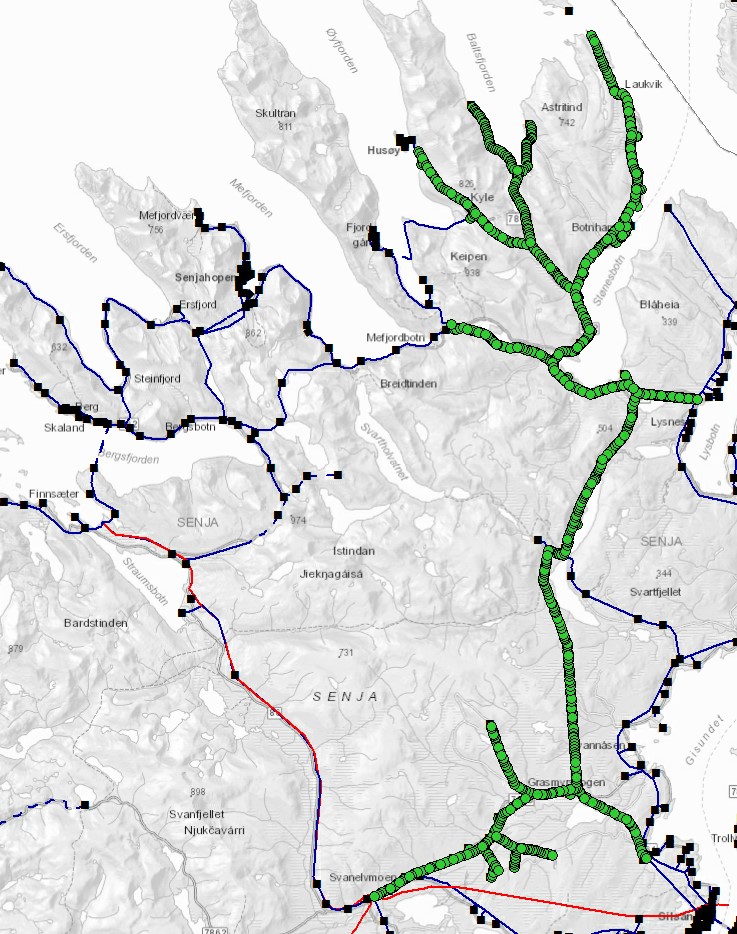}
    \caption{The SVAN22LY1 power grid. The power is distributed towards the north from the south. Each green dot represents a unique position of a utility pole.}
    \label{fig:SVAN22LY1}
\end{figure}

The power grid analyzed in this study is a radial distribution system serving an Arctic community located approximately at (69.257°N, 17.589°E). Arva Power Company, the DSO of the power grid, has named this specific grid as SVAN22LY1. Fig.~\ref{fig:SVAN22LY1} shows an overview of the whole SVAN22LY1 grid, indicated by green dots.
The SVAN22LY1 grid spans over 60 kilometers from the south to the northernmost point and has several branches to various communities towards the north. There are 978 unique utility poles (marked by green dots in Fig.~\ref{fig:SVAN22LY1}) that support the power lines. The black boxes in Fig.~\ref{fig:SVAN22LY1} represent the electric transformer stations connected to the power grids. The red lines represent a power grid with an operating voltage of 66 kV, while the blue lines represent a power grid with an operating voltage of 22 kV. The SVAN22LY1 radial grid covered by green dots has an operating voltage of 22 kV.
The largest customers connected to the SVAN22LY1 grid are located at the end of the northernmost point of the radial.

The total energy demand in the SVAN22LY1 grid is dominated by the load consumption of the local industry.
The industry performs fish processing activities that are highly seasonal and uses many electrical machines in the production line that require stable power quality. 
Even minor power disturbances in the power supply trigger significantly long interruptions since the automated production line needs to be reset. 
In particular, for every short-term power interruption that occurs, is necessary to wait from 40 minutes to 1 hour before resuming the production. 
The consequences of the power disturbances are exacerbated by the topology of the power grid, which has a radial distribution with no alternative power supply in periods with disturbances.

\section{Dataset construction}
\label{sec:dataset_construction}

\paragraph{Fault reports}
The reported faults used in this study are logged by a power-quality (PQ) metering system,
which was installed in February 2021 in the proximity of the local industries to continuously measure the power quality. 
The PQ metering system reports all incidents with a voltage variation of $\pm 10\%$ from the nominal values on each phase of a three-phase system with phases A, B, C. 
According to the standard definition, all variations of $\pm 10\%$ from normal conditions are defined as a voltage variation and a drop larger than 10\% is referred to as a voltage dip \cite{Voltage_definitions}. 
Voltage dips could provoke tripping of sensitive components such as industrial machines. 

\paragraph{Weather measurements}
The weather variables that are considered relevant in causing power faults are: wind speed of gust, wind direction, temperature, pressure, humidity, and precipitation.
The weather data are collected from areas that are more exposed to harsh weather conditions, such as hills and cliffs near the sea coast. 
To collect the weather-data in the Arctic region of interest, we used the AROME-Arctic weather model~\footnote{\url{https://www.met.no/en/projects/The-weather-model-AROME-Arctic}}. 
This model is developed by the meteorological institute of Norway (MET) and provides a reanalysis of historical weather data since November 2015 with a spatial resolution of 2.5 kilometers and a temporal resolution of 1 hour. 

To collect the weather variables, the geographical coordinates from the weather-exposed areas in the power grid are used as inputs to the AROME-Arctic model. The output from the AROME-Arctic model is a dataset of 6 weather variables from the weather-exposed areas that are analyzed.

\paragraph{Electricity load measurements}
It is reasonable to assume that some types of fault are not caused by weather phenomena but originates from external factors that influence the power flows on the grid. 
To capture these effects, 6 different power-related variables from the largest industry connected to SVAN22LY1 are collected. The variables selected as relevant to explain power faults are: difference in frequency, voltage imbalance, the difference in active and reactive power, minimum power factor, and, finally, the amount of flicker in the system. All variables are metered on three different phases (phases A, B, and C).

A \textit{change in power frequency} could be caused if there is an imbalance between energy production and consumption in the system. If there is a change in the power frequency (50 Hz is the normal frequency), the imbalance could cause power disturbances for the end-use customers. 

\textit{Voltage imbalance} is a voltage variation in the power system in which the voltage magnitudes or the phase angle between the different phases are not equal. It is believed that rapid changes (big changes within seconds/minutes) in power consumption at large industries could affect the power quality. Therefore, the \textit{difference in active and reactive power} for each phase within each minute is computed. If the difference is large, there is a high activity at the industries, which are reported by the locals to result in a larger probability for faults. 

The \textit{minimum power factor} represents the relationship between the amount of active and reactive power in the system. If the minimum power factor is low, there is an increased amount of reactive power in the system. In the end, the amount of flicker in the system is collected. 

\textit{Flicker} is considered as a phenomena in the power system and is closely connected to voltage fluctuations over a certain time frame \cite{PQ_book}. A voltage fluctuation is a regular change in voltage that happens when the machinery that requires a high load is starting. In addition, rapid changes in load demand could cause voltage fluctuations. If there are several start-up situations, or the load varies significantly during a given time frame, it will be measured a high amount of flicker in the system. The amount of flicker is particularly relevant in the industry considered in this study, as they have several large machines that require high loads and have a cyclical varying load pattern. In this study, the time frame of the flicker is 10-minutes, which is the standard for measuring the short-term flicker \cite{lovdata}. 


The PQ metering system has a 1-minute resolution, while the weather data have a 1-hour resolution.
To align the temporal resolution of the different types of variables, the power consumption data are sub-sampled by taking one sample every 60.
As an alternative sub-sampling technique, we tested taking the average of the values within each batch of 60 consecutive samples of power measurements. 
However, the results did not change significantly and, therefore, the former sub-sampling method was adopted.

\section{A brief history of explainability in deep learning}
\label{sec:nn_expl}

Due to the presence of many non-linear transformations, it is difficult to interpret the decision process of a neural network.
During the last decade, considerable research effort has been devoted towards developing insights into what a neural network learns and how it makes its decisions.
While most of the explanatory techniques were originally developed in the field of computer vision, some of them can be applied also to neural networks that process sequential or vectorial data.

Gradient based approaches aim at identifying which inputs have the most influence on the model scoring function for a given class.
The pioneering work of Simonyan at al.~\cite{simonyan2013deep} proposed to compute a saliency map by taking the gradient of the class activation score (usually, the input to the last softmax) with respect to each input features.
The visualization of the saliency maps were successively improved by using tricks such as clipping the gradients, averaging the gradients after adding Gaussian noise to the original images, and taking the absolute value of the gradients~\cite{smilkov2017smoothgrad}. 

In \cite{zeiler2014visualizing}, the authors propose a method to project the activations of an intermediate hidden layer back to the input space. The procedure consists in approximately inverting the operations of a CNN (affine transformations, ReLU activations, MaxPooling) from the hidden layer to the input layer. The result gives an insight into which details the hidden layer has captured from the input image.

The Guided Back Propagation approach performs the standard gradient back propagation but, when a ReLU is encountered, the gradient is back-propagated only if both the gradient and the ReLU activation in the forward pass are positive \cite{springenberg2014striving}.

As a drawback, gradient based methods attribute zero contribution to inputs that saturate the ReLU or MaxPool.
To capture such shortcomings, a formal notion of explainability (or relevance) was introduced in \cite{bach2015pixel}.
In particular, the authors introduced an axiom on the \textit{conservation of total relevance}, which states that the sum of relevance of all pixels must equal the class score of the model.
The authors propose to distribute the total relevance of the class score to the input features with a method called Layer-wise Relevance Propagation (LRP).
The class score is computed as the difference between the score obtained by the actual input and the score obtained by an uninformative input, called \textit{baseline}.
Each time the relevance is passed down from a neuron to the contributing neurons in the layer below, the total relevance of contributing neurons is preserved.
All incoming relevances to a neuron from the layer above are collected and summed up before passing down further. By doing this recursively from layer to layer, the input layer is eventually reached, which gives the relevance of each input feature.
The relevance of a neuron to its contributing inputs can be distributed based on the magnitude of the weights of the neural network layers.

While LRP followed the conservation axiom, it did not formalize how to distribute the relevance among the input features. To address this problem DeepLiFT~\cite{shrikumar2017learning} enforces an additional axiom on how to propagate the relevance down, by following the chain rule like gradients.

